\documentclass[letterpaper]{article} 
\usepackage{aaai25}  
\usepackage{times}  
\usepackage{helvet}  
\usepackage{courier}  
\usepackage[hyphens]{url}  
\usepackage{graphicx} 
\usepackage{natbib}  
\usepackage{caption} 
\usepackage{algorithm}
\usepackage{algorithmic}

\usepackage{newfloat}
\usepackage{listings}
\DeclareCaptionStyle{ruled}{labelfont=normalfont,labelsep=colon,strut=off} 
\floatstyle{ruled}
\newfloat{listing}{tb}{lst}{}
\floatname{listing}{Listing}
\title{Bridging Traffic State and Trajectory for Dynamic Road Network and Trajectory Representation Learning}
\author{
    Chengkai Han\textsuperscript{\rm 1}, Jingyuan Wang\textsuperscript{\rm 1,2,3}\thanks{Corresponding author: jywang@buaa.edu.cn}, Yongyao Wang\textsuperscript{\rm 1}, Xie Yu\textsuperscript{\rm 1}, Hao Lin\textsuperscript{\rm 2,3}, Chao Li\textsuperscript{\rm 1,4}, Junjie Wu\textsuperscript{\rm 2,3}
}
\affiliations{
    \textsuperscript{\rm 1} School of Computer Science and Engineering, Beihang University, Beijing, China\\
    \textsuperscript{\rm 2} MIIT Key Laboratory of Data Intelligence and Management, Beihang University, Beijing, China\\
    \textsuperscript{\rm 3} School of Economics and Management, Beihang University, Beijing, China\\
    \textsuperscript{\rm 4} Shenzhen Institute of Beihang University, Shenzhen, China
}

\usepackage{bibentry}

\usepackage{xspace}
\usepackage{bm}
\usepackage{amsmath}
\usepackage{amsthm}
\usepackage{amsfonts}
\usepackage{booktabs}
\newtheorem{mydef}{Definition}
\newcommand{\ie}{\emph{i.e.,}\xspace}
\newcommand{\eg}{\emph{e.g.,}\xspace}
\newcommand{\name}{TRACK\xspace}
\newcommand{\paratitle}[1]{\vspace{1.5ex}\noindent\textbf{#1}}

\begin{document}

\maketitle

\begin{abstract}
Effective urban traffic management is vital for sustainable city development, relying on intelligent systems with machine learning tasks such as traffic flow prediction and travel time estimation. Traditional approaches usually focus on static road network and trajectory representation learning, and overlook the dynamic nature of traffic states and trajectories, which is crucial for downstream tasks. To address this gap, we propose TRACK, a novel framework to bridge traffic state and trajectory data for dynamic road network and trajectory representation learning.
TRACK leverages graph attention networks (GAT) to encode static and spatial road segment features, and introduces a transformer-based model for trajectory representation learning. By incorporating transition probabilities from trajectory data into GAT attention weights, TRACK captures dynamic spatial features of road segments. Meanwhile, TRACK designs a traffic transformer encoder to capture the spatial-temporal dynamics of road segments from traffic state data.
To further enhance dynamic representations, TRACK proposes a co-attentional transformer encoder and a trajectory-traffic state matching task. Extensive experiments on real-life urban traffic datasets demonstrate the superiority of TRACK over state-of-the-art baselines. Case studies confirm TRACK's ability to capture spatial-temporal dynamics effectively. 
\end{abstract}

\begin{links}
     \link{Code}{https://github.com/NickHan-cs/TRACK}
\end{links}


\section{Introduction}
Intelligent urban traffic management, such as traffic flow prediction~\cite{AGCRN,wang2022traffic, ji2022stden,liu2024full}, travel time estimation~\cite{DBLP:conf/aaai/WangZCLZ18,DBLP:conf/cikm/WuWZJ19} and trajectory analysis~\cite{chen2015efficient, ding2018ultraman, chen2019real}, plays a crucial role in ensuring efficient city functioning and promoting sustainable development~\cite{LibCity,survey}. In the realm of intelligent urban traffic management, traffic state data and trajectory data are two core components that can encapsulate the macroscopic and microscopic characteristics of cities, respectively, and their representation learning, \ie learning generic low-dimensional road segment and trajectory vectors, serve as two fundamental pillars for various urban traffic tasks~\cite{chen2018price,wang2019empowering,DBLP:conf/aaai/JiangZWJ23}.

\begin{figure}[t!]
    \centering
    \includegraphics[width=0.7\columnwidth]{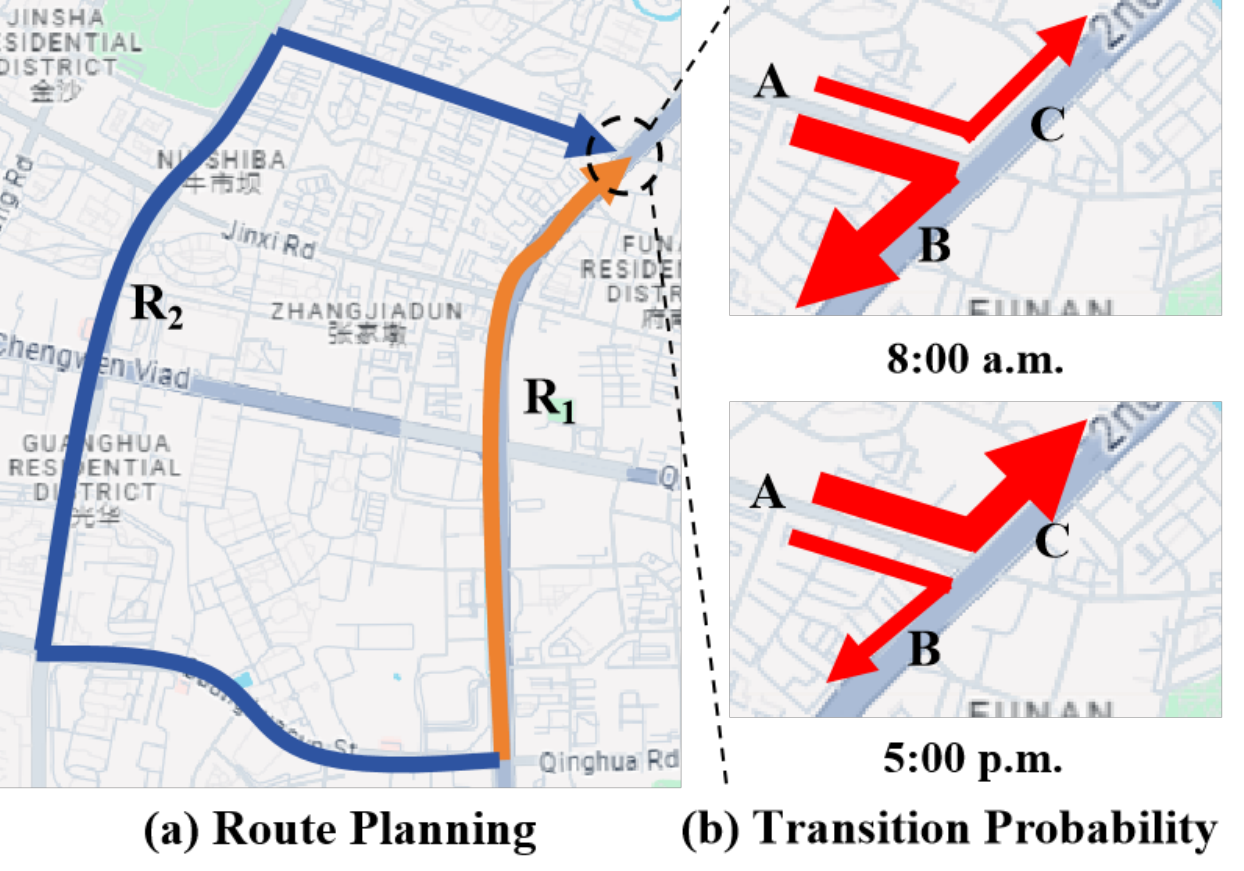}
    \caption{An example of mutual influences between traffic state data and trajectory data.
    }\label{fig:intro}
\end{figure}

Recently, many efforts have been devoted to modeling traffic state data~\cite{PDFormer, MultiSPANS, CMuST, ji2022precision, ji2023spatio,ji2020interpretable} and trajectory data~\cite{PIM,Trembr,START}. However, existing methods typically model these two types of data independently, lacking approaches capable of jointly modeling them. In urban transportation scenarios, traffic state data describes the dynamic macroscopic characteristics of groups on the road network, while trajectory data reflects the dynamic movement attributes of individuals on the road network. There are spatio-temporal correlations and mutual influences between traffic state data and trajectory data.

Firstly, traffic states influence individuals' choices of the trajectory route. For example, as shown in Figure~\ref{fig:intro}(a), people tend to choose the shortest route $R_1$ during non-peak hours, but it might be a better choice to detour with a more time-saving route $R_2$ during peak hours (\ie when the average traffic speed of $R_1$ is low). Traffic states naturally affect the travel time of each road segment in a trajectory. Secondly, individual transitions on the road network are the direct cause of changes in traffic states. As shown in Figure (b), the traffic flow on segment C consists of the traffic flow entering from segment A and that entering from segment B. Therefore, the transition probability from segment A to segment C varies over different time periods, directly impacting the traffic states on segment C. Therefore, jointly modeling traffic state data and trajectory data can enrich the spatio-temporal information learned by the model, capturing the dynamic nature of traffic states and trajectories.

To achieve this, we propose to bridge two prominent types of dynamic data, \ie \underline{TRA}ffic state and trajectory, for dynami\underline{C} road networ\underline{K} and trajectory representation learning (\name). Specifically, \name encodes the static and spatial features of road segments with graph attention networks (GAT) to learn road segment representations, followed by a trajectory transformer encoder with the masked trajectory prediction task and the contrastive trajectory learning task to learn trajectory representations. To capture the dynamic spatial features of road segments, we incorporate the transition probabilities computed from the trajectory data into the attention weights of GAT. Meanwhile, to capture the spatial-temporal dynamics of road segments from traffic state data, we learn a traffic transformer encoder with the mask state prediction task and the next state prediction task. More importantly, to capture the interactions of traffic state and trajectory data in characterizing a road segment's dynamic features, we model the information exchange between different data views by designing a co-attentional transformer encoder with a novel gravitivity-based attention mechanism and the trajectory-traffic state matching task. Finally, we pre-train the whole model with a joint self-supervised learning loss.

We conduct extensive experiments on two real-life urban traffic datasets and compare our proposed \name method with several state-of-the-art baselines. Evaluated with two downstream tasks, \name achieves consistently superior performances over baselines. Case studies also validate that \name can capture the spatial-temporal dynamics of road segments and trajectories through learned dynamic representations.

\section{Preliminaries}

In this section, we introduce the mathematical notations used throughout the paper and formally define our research problem.

\subsection{Basic Elements of Urban Traffic}
We start by introducing the basic spatial-temporal units of urban traffic, \ie a road segment and a time slice.
\begin{mydef}[\textbf{Road Segment}]
A road segment $v \in \mathcal{V}$ is the minimum spatial unit in urban traffic scenarios where $\mathcal{V}$ is the set of road segments.
\end{mydef}

\begin{mydef}[\textbf{Time Slice}]
 A time slice $t$ is the minimum time unit (\eg an hour) in urban traffic scenarios.
\end{mydef}

For convenience, we might call \textit{segment} for short in unambiguous cases. Next, we characterize the \textbf{spatial}, \textbf{static}, \textbf{dynamic} features of road segments and the \textbf{trajectory} with the following concepts.
\begin{mydef}[\textbf{Road Network}]
A road network is characterized as a graph $\mathcal{G}=(\mathcal{V},{\bm A})$, where $\mathcal{V}=\{v_1,\cdots,v_N\}$ is a node set of $N$ road segments and ${\bm A} \in \mathbb{R}^{N \times N} $ is an adjacency matrix to capture the link information between $N$ road segments. The road network entails the spatial features of road segments.
\end{mydef}

\begin{mydef}[\textbf{Static Feature of Road Segment}]
The static feature $\boldsymbol{f}_v \in \mathbb{R}^{C_1}$ for a road segment $v$ is a feature vector with which $v$ is generally associated and does not change over time after it was built. $C_1$ is the dimension of the feature vector. For example, $C_1=5$ if the features include longitude, latitude, segment type, length, and speed limit.
\end{mydef}

\begin{mydef}[\textbf{Traffic State Sequence}]
A traffic state sequence $\mathcal{S}_t \in \mathbb{R}^{T \times N \times C_2}$ is composed of $T$ consecutive historical traffic states before the time slice $t$, where $\mathcal{S}_t=(\bm{TS}_{t-T},\cdots,\bm{TS}_{t-1})$. $\bm{TS}_{t}$ denotes a traffic state at the time slice $t$. A traffic state $\bm{TS}_t \in \mathbb{R}^{N \times C_2}$ is the statistics (\eg flow, density, average speed) on $N$ road segments within the time slice $t$, where $C_2$ is the dimension of the statistics. The traffic state sequence involves the dynamic features of road segments that will change over time. 
\end{mydef}

\begin{mydef}[\textbf{Trajectory}]
A trajectory $\mathcal{T}=[\langle v_i,t_i\rangle]_{i=1}^m$ is a sequence of spatial-temporal points that record the movement behavior of a car or person in the scope of the road network $\mathcal{G}$, where $m$ is the total number of spatial-temporal points, $v_i \in \mathcal{V}$ denotes the segment for the $i$-th visit, and $t_i \in \mathbb{R}$ denotes the corresponding visit timestamp.
\end{mydef}

We use $\bm{F}_{\mathcal{V}} \in \mathbb{R}^{N \times C_1}$ to denote the static feature matrix of $N$ road segments in the road network $\mathcal{G}$. We assume that there is a set of $k_D$ trajectories within the time slice $t$, denoted as a trajectory set $\mathcal{D}_t=\{\mathcal{T}_j\}_{j=1}^{k_D}$, indicating that the departure timestamp of each trajectory $\mathcal{T}_j \in \mathcal{D}_t$ lies in the time slice $t$.

\subsection{Problem Formulation}
We formulate two representation learning tasks in urban traffic scenarios, \ie \textit{Dynamic Road Network Representation Learning (DRNRL)} and \textit{Trajectory Representation Learning (TRL)}.

\begin{mydef}[\textbf{Dynamic Road Network Representation Learning}]
Given the road network $\mathcal{G}$, the historical traffic state sequence $\mathcal{S}_t$ and the trajectory set $\mathcal{D}_t$ at the time slice $t$, \textit{DRNRL} aims to derive a generic $d_s$-dimensional representation $\boldsymbol{h}_{v,t} \in \mathbb{R}^{d_s}$ at the time slice $t$ for each road segment $v \in \mathcal{V}$ on the road network. 
\end{mydef}

\begin{mydef}[\textbf{Trajectory Representation Learning}]
Given the road network $\mathcal{G}$, the historical traffic state sequence $\mathcal{S}_t$ and the trajectory set $\mathcal{D}_t$ at the time slice $t$, \textit{TRL} aims to derive a generic $d_t$-dimensional representation $\boldsymbol{l}_{\mathcal{T}} \in \mathbb{R}^{d_t}$ for each trajectory $\mathcal{T} \in \mathcal{D}_t$.
\end{mydef}

Traditional representation learning methods on road network~\cite{HRNR} usually focus on learning a road segment's representation that does not change over time. However, in great contrast, \textit{DRNRL} aims to learn dynamic representations of road segments by considering the dynamic features derived from traffic state and trajectory data. We interchangeably use the terms \textit{road network representation} and \textit{road segment representation} hereinafter. We assume that the numbers of latent dimensions for \textit{DRNRL} and \textit{TRL} are set to the same value $d$ in our problem, \ie $d=d_s=d_t$. The learned road segment representation can be applied to various segment-related downstream tasks such as traffic state prediction and on-demand service prediction. The learned trajectory representation can be applied to various trajectory-related downstream tasks such as travel time estimation
and anomalous trajectory detection. 

\section{Methodology}

\begin{figure}[bt!]
    \centering
    \includegraphics[width=1\columnwidth]{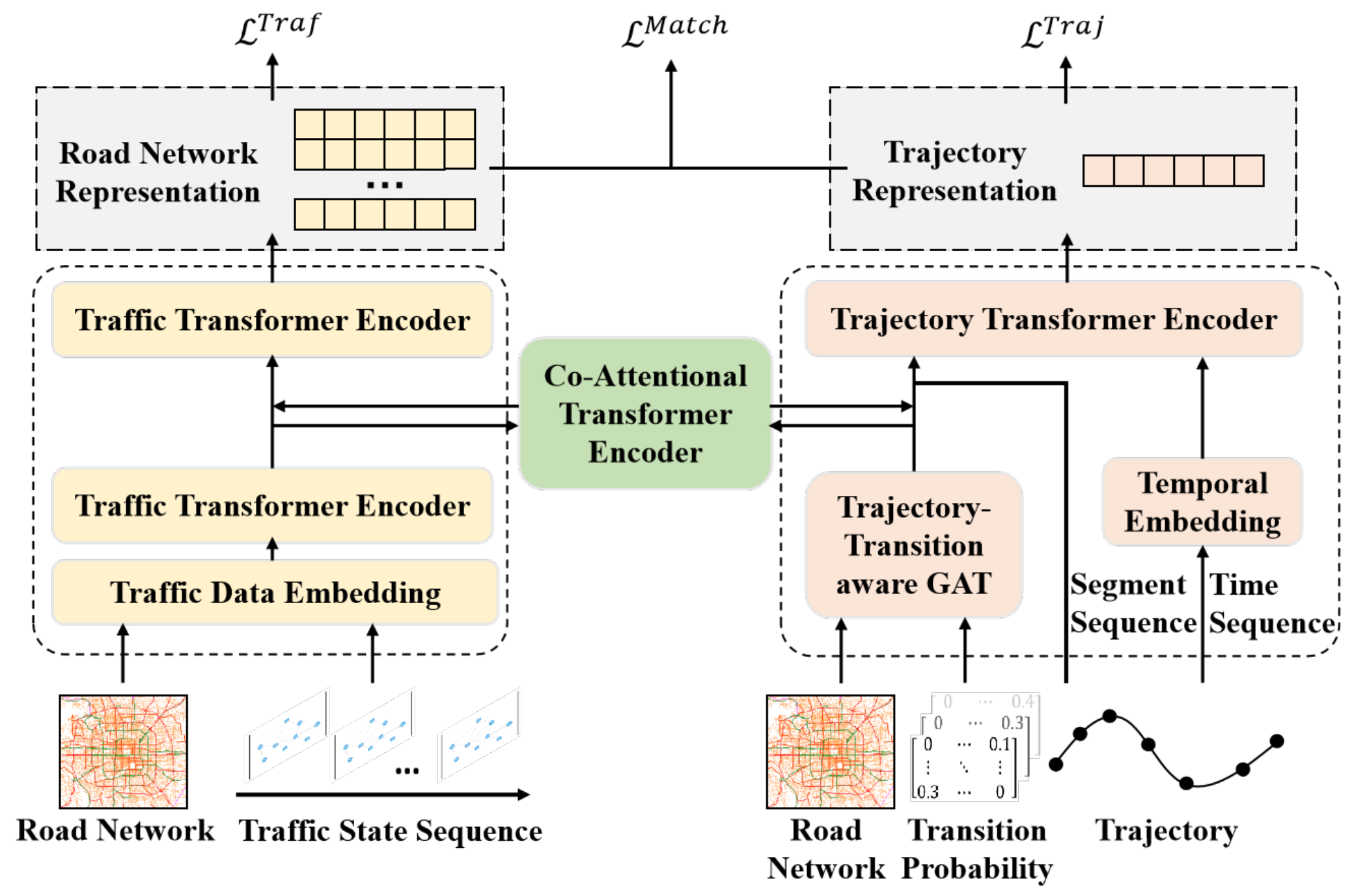}
    \caption{The overall architecture of the \name model.}
    \label{fig:TRACK}
\end{figure}

In this section, we present the proposed \textit{\name} model. Our core idea is to incorporate dynamic information from traffic state and trajectory data into road network and trajectory representation learning, and model the information exchange between multi-view data to enhance the dynamic representations. The overall architecture of the proposed model is shown in Figure~\ref{fig:TRACK}.

\subsection{Basic Pipeline of TRL}\label{sec:BPTRL}
In this part, we introduce a basic pipeline for TRL, which first encodes the segments appearing in a trajectory into low-dimensional vectors and then combines them with the timestamp representations to derive the trajectory's final representation vector. We train it with the Masked Trajectory Prediction (MTP) task and the Contrastive Trajectory Learning (CTL) task. 

\paratitle{Encoding Road Segment's Static and Spatial Features}\label{sec:rnss}.
We aim to project each segment $v$ in the road network $\mathcal{G}$ into a low-dimensional representation vector $\boldsymbol{h}^{\mathcal{G}}_{v} \in \mathbb{R}^{d}$. The static features of each segment, \eg length and speed limits, contain rich semantics of the segment. Moreover, the connectivity of the segments in the road network, \ie the local network structure, also entails the spatial semantics of a segment. To this end, it is natural to learn the representation vector of a segment from both its static features and the local network structure. We adopt a GNN method, \ie a multi-layer Graph Attention Network (GAT)\cite{GAT}, to model each segment's static and spatial features as follows:
\begin{equation}
    {\bm H}^{Traj} = \mathrm{GAT}({\bm F_{\mathcal{V}}}, {\bm A}),
\end{equation}
where $\mathrm{GAT}(\cdot,\cdot)$ denotes the implementation of a standardized GAT or a GAT optimized with sparse matrix operations, and ${\bm H}^{Traj} = [\boldsymbol{h}^{Traj}_{v}]^{N}_{v=1} \in \mathbb{R}^{N \times d}$ is a matrix form of $N$ segments's representations in $\mathcal{G}$.

\paratitle{Encoding Timestamp Information.} 
We introduce a temporal embedding layer to transform raw timestamps in trajectories into low-dimensional representation vectors. Specifically, it contains $\boldsymbol{t}^{weekly}_{i},\boldsymbol{t}^{daily}_{i},\boldsymbol{t}^{pos}_i,\boldsymbol{t}^{interval}_{i}\in\mathbb{R}^{d}$, which represent weekly periodic patterns, daily periodic patterns, position information, time interval information, respectively.

\paratitle{Encoding the Whole Trajectory}\label{sec:ETD}.
The whole trajectory can be divided into a segment sequence and a time sequence. Therefore, for the $i$-th visit, we first feed the segment $v_i$ into the GAT to obtain the segment representation $\boldsymbol{h}^{Traj}_{v_i}$. Meanwhile, we feed the time sequence into the temporal embedding layer to obtain $\boldsymbol{t}^{weekly}_{i}$, $\boldsymbol{t}^{daily}_{i}$, $\boldsymbol{t}^{pos}_{i}$ and $\boldsymbol{t}^{interval}_{i}$. Then, we derive the overall representation $\boldsymbol{l}_i \in \mathbb{R}^{d}$ for the $i$-th visit in a trajectory by as follows:
\begin{equation}
    \boldsymbol{l}_i = \boldsymbol{h}^{Traj}_{v_i} + \boldsymbol{t}^{weekly}_{i} + \boldsymbol{t}^{daily}_{i} + \boldsymbol{t}^{pos}_{i} + \boldsymbol{t}^{interval}_{i}.
\end{equation}
In order to capture the long-range dependencies of visits in a trajectory and identify the global semantics of the trajectory, we further feed the representation sequence $[\boldsymbol{l}_i]_{i=1}^{m}$ into a transformer encoder~\cite{Transformer} to obtain the final trajectory representation $\boldsymbol{l}_{\mathcal{T}}\in\mathbb{R}^{d}$
which can be mathematically defined as follows:
\begin{equation}
    \boldsymbol{l}_{\mathcal{T}} = \mathrm{TrajTrans}(\boldsymbol{ph},\boldsymbol{l}_1,\cdots,\boldsymbol{l}_m)[0], 
\end{equation}
where the function $\mathrm{TrajTrans}(\cdot)$ denotes a standard transformer encoder or a transformer-like encoder and $\boldsymbol{ph}$ is the embedding vector of the placeholder.

\paratitle{Masked Trajectory Prediction}. The general idea of this task is to mask the consecutive subsegments of the trajectory and their corresponding timestamps, and to use linear layers to predict the masked values, \ie $\hat{y}^{S} \in \mathbb{R}^{|\mathcal{T}| \times |\mathcal{V}|}$ and $\hat{y}^{T} \in \mathbb{R}^{|\mathcal{T}|}$, respectively. The loss functions can be defined as follows:
\begin{equation}
    \mathcal{L}^{Traj}_{S} = -\frac{1}{|\mathcal{M}_S|}\sum_{v_i\in\mathcal{M}_S}\mathrm{log}\frac{\mathrm{exp}(\hat{y}^{S}_{v_i})}{\sum_{v_j\in\mathcal{V}}\mathrm{exp}(\hat{y}^{S}_{v_j})},
\end{equation}
\begin{equation}
    \mathcal{L}^{Traj}_{T} = \frac{1}{|\mathcal{M}_T|}\sum_{t_i\in \mathcal{M}_T}|\hat{y}^T_{v_i} - t_i|,
\end{equation}
where $\mathcal{M}_S$ and $\mathcal{M}_T$ denote the sets of masked road segments and masked timestamps, respectively.

\paratitle{Contrastive Trajectory Learning}. The general idea of this task is to adopt a contrastive learning strategy to generate multiple samples of a trajectory from different views and bring semantically similar samples closer in the representation space while dispersing dissimilar samples. The loss function can be defined as follows:
\begin{equation}
     \mathcal{L}^{Traj}_{con}(\mathcal{T}_i, \mathcal{T}_j) = -\log\frac{\exp(\mathrm{sim}(\boldsymbol{l}_{\mathcal{T}_i}, \boldsymbol{l}_{\mathcal{T}_j})/\tau)}{\sum_{k=1}^{2B}\bm{1}_{[k\neq i]}\exp(\mathrm{sim}(\boldsymbol{l}_{\mathcal{T}_i}, \boldsymbol{l}_{\mathcal{T}_k})/\tau)},
\end{equation}
where $B$ is the batch size, $(\mathcal{T}_i, \mathcal{T}_j)$ is a positive pair in the batch, $\bm{1}_{[k\neq i]}\in\{0, 1\}$ is an indicator function that is equal to 1 if condition $k\neq i$ is satisfied and $\tau$ denotes a temperature parameter. The overall loss of this task for the batch, \ie $\mathcal{L}^{Traj}_{con}$, is computed by averaging the losses of all positive pairs in the batch.

\subsection{Modeling Road Segments' Dynamic Features}
\label{sec:rnrl}
In this part, we model the dynamic features of road segments that can change over time, including the spatial-temporal dynamics in trajectories and traffic state sequences.

\paratitle{Encoding Road Segment's Dynamic Spatial Features with Trajectory Data}.
Trajectory data entails some dynamic spatial semantics of segments. For example, a large transition probability between two segments revealed from the trajectory data may indicate that the two segments are nearby in the semantic space. Therefore, we replace the standard GAT in Section \textit{Encoding Road Segment’s Static and Spatial Features} with a Trajectory Transition-aware GAT. Specifically, based on the trajectories, we introduce a time-aware transition probability $p_{i,j,t}$ for the time slice $t$ to capture the transition patterns between two segments $v_i$ and $v_j$. $p_{i,j,t}$ considers the historical trajectories that occur periodically within the time slice $t$. To incorporate $p_{i,j,t}$ into the GAT, we compute a normalized attention weight $\alpha_{i,j,t}$ between $v_i$ and $v_j$ as follows:
\begin{equation}
     \alpha_{i,j,t}=\frac{\exp(\mathrm{LeakyReLU}(e_{i,j,t}))}{\sum_{k\in\mathcal{N}_{v_i}}\exp(\mathrm{LeakyReLU}(e_{i,k,t}))},
\end{equation}
\begin{equation}
     e_{i,j,t}=(\boldsymbol{h}_{v_i,t}^{\prime}{\bm W}_1+\boldsymbol{h}_{v_j,t}^{\prime}{\bm W}_2+p_{i,j,t}{\bm W}_3){\bm W}_4^\top,
\end{equation}
where ${\bm W}_1,{\bm W}_2\in\mathbb{R}^{d\times d'}$ and ${\bm W}_3,{\bm W}_4\in\mathbb{R}^{1\times d'}$ are learnable weight parameters, $\mathcal{N}_{v_i}$ is the neighborhood set of road segment $v_i$.

\paratitle{Traffic Data Embedding.} The Traffic Data Embedding layer is designed to convert the traffic state sequence at time slice $t$, \ie $\mathcal{S}_t \in \mathbb{R}^{T \times N \times C_2}$, into an embedding tensor, \ie $\mathcal{X}^{Traf}_t \in \mathbb{R}^{T \times N \times d_x}$. Specifically, we first project $\mathcal{S}_t$ directly into a $d_x$-dimensional representation tensor, \ie $\mathcal{X}^{raw}_t = FC(\mathcal{S}_t) \in \mathbb{R}^{T\times N\times d_x}$, through a fully-connected feed-forward network $FC(\cdot): \mathbb{R}^{T\times N\times C_{2}} \rightarrow \mathbb{R}^{T\times N\times d_x}$. Next, we employ three temporal embeddings, \ie ${\bm X}^{weekly}_{t},{\bm X}^{daily}_{t},{\bm X}^{pos}_t\in\mathbb{R}^{T\times d_x}$, to extract the weekly periodic patterns, daily periodic patterns and position information for all $T$ time slices, respectively. To further model the spatial information of the road network, we also employ a multi-layer GAT to generate an embedding matrix $\bm{X}^{\mathcal{G}}\in\mathbb{R}^{N\times d_x}$ of the road network. The final embedding tensor $\mathcal{X}^{Traf}_t$ can then be computed as follows:
\begin{equation}
     \mathcal{X}^{Traf}_{t}=\mathcal{X}^{raw}_t + {\bm X}^{weekly}_{t} + {\bm X}^{daily}_{t} + {\bm X}^{pos}_t + {\bm X}^{\mathcal{G}}.
\end{equation}

\paratitle{Traffic Transformer Encoder.} 
We further feed $\mathcal{X}^{Traf}_{t}$ into a traffic transformer encoder to model dynamic spatial-temporal dependencies hidden in traffic state sequences. The traffic transformer encoder is composed of multiple traffic transformer encoder layers. In each encoder layer, we first feed $\mathcal{X}^{Traf}_{t}$ into a spatial encoder and a temporal encoder, respectively. The spatial encoder takes a geographical and semantic neighbor-aware GAT layer to capture the dynamic spatial dependencies of traffic states in each time slice, whereas the temporal encoder takes a temporal self-attention layer to capture the dynamic temporal patterns for different road segments in the traffic state data. Next, we concatenate the output embedding of the spatial and temporal encoders to form a fusion representation, which is further fed into other components of a transformer encoder, \eg Add \& Norm layer and Feed Forward layer. At the last of the traffic transformer encoder, we use a convolutional layer to transform the output embedding tensor into a matrix ${\bm H}^{Traf}_t\in\mathbb{R}^{N\times d}$, which is the final representation for $N$ road segments at the time slice $t$. We summarize the computation of ${\bm H}^{Traf}_t$ as follows:
\begin{equation}\label{eq:traf_rep}
     {\bm H}^{Traf}_t=\mathrm{TrafTrans}(\mathcal{X}^{Traf}_{t}),
\end{equation}
where the function $\mathrm{TrafTrans}(\cdot): \mathbb{R}^{T \times N \times d_x} \rightarrow \mathbb{R}^{N \times d}$ denotes the whole Traffic Transformer Encoder. Actually, other feasible spatio-temporal encoders can also replace this encoder.

\paratitle{Pre-training Traffic Data Embedding and Traffic Transformer Encoder via Mask State Prediction and Next State Prediction.} We design two self-supervised tasks to learn generic segment representations. Specifically, we randomly mask a sequence of historical traffic states for each segment and use the generated intermediate representations to predict the masked traffic states, while using the generated segment representations of the next time slice to predict the traffic state of the next time slice. Ultimately, the loss $\mathcal{L}^{Traf}$ is obtained by the weighted sum of the losses from two tasks.

\subsection{Modeling Multi-view Information Exchange}

We propose a co-attentional transformer encoder to tackle the multi-view information exchange between different data modality, followed by a trajectory-traffic state matching task to maximize the consistency between segment representations and trajectory representations.

\begin{figure}[t]
    \centering
    \includegraphics[width=1\columnwidth]{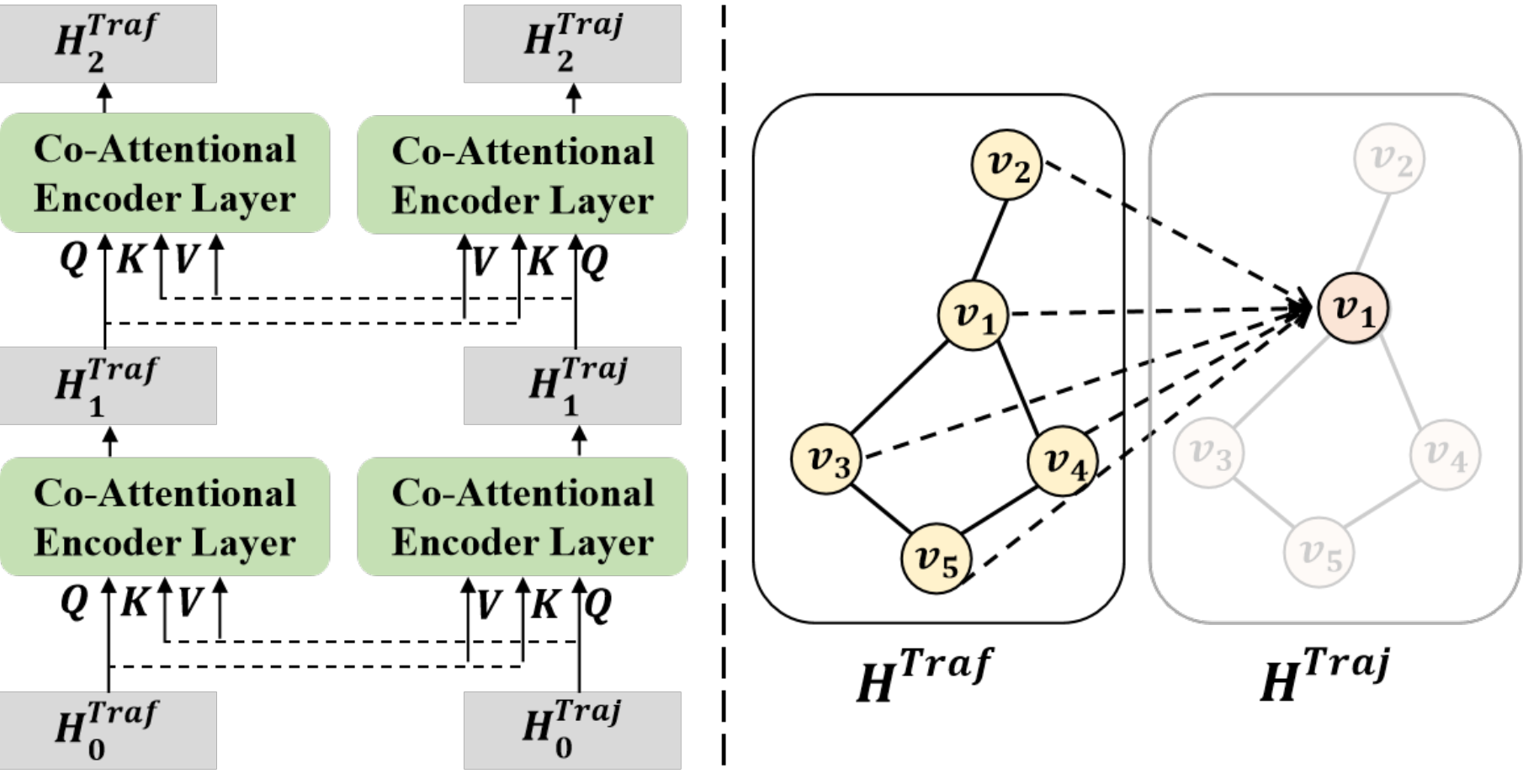}
    \caption{Framework of the Co-Attentional Transformer Encoder and the GAT Operation from the View of Trajectory Data.}
    \label{fig:cogat}
\end{figure}

\paratitle{Co-Attentional Transformer Encoder.} In each view, the key idea for the co-attentional transformer encoder is to replace the transformer encoder's multi-head self-attention module with a GAT layer, as shown in Figure~\ref{fig:cogat}, which aggregates the neighborhood nodes' features from the other view to form the node's feature representation in the target view. Moreover, in the GAT layer, inspired by Newton’s law of universal gravitation~\cite{gravity}, we assume that the influence between two segments decreases with the distance between them and design a novel way to compute the attention weight.

Specifically, we take the view of trajectory data to describe the mechanism of the co-attentional transformer encoder, and the co-attentional transformer encoder for the other view works similarly. We first define a $K$-minute reachable neighborhood set for segment $v_i$ as $\mathcal{R}_{v_i}$, which is the set of segments that can reach $v_i$ in $K$ minutes. We use $\boldsymbol{h}^{Traj}_{v_i} \in \mathbb{R}^{d}$ to denote the representation vector of segment $v_i$ at the time slice $t$ produced by Trajectory Transition-aware GAT. To model the influence of information from the other view over $\boldsymbol{h}^{Traj}_{v_i}$, we employ a GAT layer, which defines a normalized attention weight $\alpha^{Traj}_{i,j}$ between $v_i$ and $v_j \in \mathcal{R}_{v_i}$ as follows:
\begin{equation}
\alpha^{Traj}_{i,j}=\frac{\exp(\mathrm{LeakyReLU}(e^{Traj}_{i,j}))}{\sum_{k\in\mathcal{R}_{v_i}}\exp(\mathrm{LeakyReLU}(e^{Traj}_{i,k}))},
\end{equation}
\begin{equation}\label{eq:att_GATGravity}
e^{Traj}_{i,j}=(\boldsymbol{h}^{Traj}_{v_i}{\bm W}_5+\boldsymbol{h}^{Traf}_{v_j}{\bm W}_6+\mathrm{deter}(v_i,v_j){\bm W}_7){\bm W}_8^\top,
\end{equation}
where ${\bm W}_5,{\bm W}_6\in\mathbb{R}^{d\times d'}$ and ${\bm W}_7, {\bm W}_8\in\mathbb{R}^{1\times d'}$ are learnable parameters,
the function $\mathrm{deter}(\cdot)$ denotes a geographical distance deterrence function such as Negative Power Function, and $\boldsymbol{h}^{Traf}_{v_j}$ denotes the representation vector of segment $v_j \in \mathbb{R}^{d}$ at the time slice $t$ produced by the Traffic Transformer Encoder in the other view. The GAT layer uses the normalized attention weight $\alpha^{Traj}_{i,j}$ to aggregate the features of neighboring segments in the other view to obtain the feature representation of the targeted segment. The rest of the transformer block proceeds as before. 


\paratitle{Trajectory-Traffic State Matching.} We design a contrastive learning task, \ie \textit{Trajectory-Traffic State Matching}, which maximizes the agreement of a trajectory's representation vector and the corresponding road segment sequence's representation vector generated from traffic state data. The core of this task is that the trajectories of the current time slice correspond to the traffic state of the current time slice, rather than traffic states of other time slices.

Specifically, for a trajectory $\mathcal{T}=[\langle v_i,t_i\rangle]_{i=1}^m$, its representation $\boldsymbol{l}_{\mathcal{T}}$ can be extracted by the TRL process. In the meanwhile, a representation matrix ${\bm RS}^{Traf}_{\mathcal{T}} \in \mathbb{R}^{m \times d}$ for the trajectory's corresponding segment sequence $[v_i]_{i=1}^m$ can be extracted based on the traffic state data as follows:
\begin{equation}
   {\bm RS}^{Traf}_{\mathcal{T}} = [\boldsymbol{h}^{Traf}_{v_i,t}]^{m}_{i=1},
\end{equation}
where $\boldsymbol{h}^{Traf}_{v_i,t} \in \mathbb{R}^{d}$ denotes the representation vector for road segment $v_i$ at the timestamp $t_1$ (within the time slice $t$), and is derived based on ${\bm H}^{Traf}_t$ as described in Equation~(\ref{eq:traf_rep}). Then, we can obtain the final representation vector $\boldsymbol{h}^{Traf}_{\mathcal{T},t} \in \mathbb{R}^{d}$ by average pooling over the first dimension of the matrix ${\bm RS}^{Traf}_{\mathcal{T}}$.

In the contrastive learning process, We define the spatio-temporal data of $B_t$ time slices $\Theta$ as a training batch. For each trajectory $\mathcal{T}\in\mathcal{D}_t$ within the time slice $t\in\Theta$, we regard $(\boldsymbol{l}_{\mathcal{T}}, \boldsymbol{h}^{Traf}_{\mathcal{T}})$ as a positive pair. To construct negative pairs, we obtain $\boldsymbol{h}^{Traf}_{\mathcal{T},t^{\prime}} \in \mathbb{R}^{d}$ from other time slices $t^{\prime}\in\Theta$. Then, a negative pair can be constructed as $(\boldsymbol{l}_{\mathcal{T}}, \boldsymbol{h}^{Traf}_{\mathcal{T},t^{\prime}})$. Formally, the loss function of the positive pair $(\boldsymbol{l}_{\mathcal{T}}, \boldsymbol{h}^{Traf}_{\mathcal{T},t})$ for the contrastive learning based on NT-Xent is defined as:
\begin{equation}
    d_{\mathcal{T},t}=\exp(\mathrm{sim}(\boldsymbol{l}_{\mathcal{T}}, \boldsymbol{h}^{Traf}_{\mathcal{T},t})/\tau),
\end{equation}
\begin{equation}
    \mathcal{L}^{Match}(\boldsymbol{l}_{\mathcal{T}}, \boldsymbol{h}^{Traf}_{\mathcal{T},t})=-\log\frac{d_{\mathcal{T},t}}{d_{\mathcal{T},t}+\sum_{t^{\prime} \in \Theta}d_{\mathcal{T},t^{\prime}}},
\end{equation}
where $\tau$ denotes a temperature parameter. The total loss of contrastive learning for the batch, \ie $\mathcal{L}^{Match}$, is calculated by averaging the losses of all positive pairs in the batch.

\subsection{Joint Pre-training for the Whole Model}

To facilitate learning spatial-temporal patterns across multi-source data, we pre-train all modules of the proposed model in a joint manner, which defines the following loss function:
\begin{equation}
\mathcal{L}=\lambda^{Traj}\mathcal{L}^{Traj}+\lambda^{Traf}\mathcal{L}^{Traf}+\lambda^{Match}\mathcal{L}^{Match},
\end{equation}
where $\lambda^{Traj},\lambda^{Traf},\lambda^{Match}$ are three hyper-parameters that control the influence of each individual loss function over the proposed model, respectively.

\section{Experiments}

\subsection{Experimental Setup}

\begin{table}[bt!]
  \centering
  
    \begin{tabular}{cccc}
    \toprule
    Dataset & \#Road Segment & \#Edge & \#Trajectory \\
    \midrule
    Xi'an & 5,168  & 12,643 & 834,560 \\
    Chengdu & 6,153  & 15,783 & 1,262,406 \\
    \bottomrule
    \end{tabular}%
\caption{Statistics of the Two Datasets.}
  \label{tab:data_des}%
\end{table}%

\begin{table*}[t]
  \centering
  
  \resizebox{2\columnwidth}{!}{
    \begin{tabular}{c|ccc|ccc||c|ccc|ccc}
    \toprule
    \multicolumn{7}{c||}{Multi-Step Traffic State Prediction} & \multicolumn{7}{c}{Travel Time Estimation} \\
    \midrule
    Dataset & \multicolumn{3}{c|}{Xi'an} & \multicolumn{3}{c||}{Chengdu} & Dataset & \multicolumn{3}{c|}{Xi'an} & \multicolumn{3}{c}{Chengdu} \\
    \midrule
    Models & MAE   & MAPE(\%) & RMSE  & MAE   & MAPE(\%) & RMSE  & Models & MAE   & MAPE(\%) & RMSE  & MAE   & MAPE(\%) & RMSE \\
    \midrule
    DCRNN & 1.288 & 16.38 & 2.491 & 1.554 & 18.21 & 2.860 & traj2vec & 1.667 & 23.44 & 2.465 & 1.296 & 22.16 & 1.915 \\
    GWNET & 1.297 & 15.58 & 2.334 & 1.610 & 18.13 & 2.707 & t2vec & 1.663 & 23.38 & 2.463 & 1.296 & 22.21 & 1.923 \\
    MTGNN & 1.222 & 14.90 & \underline{2.161} & 1.470 & \underline{16.76} & 2.495 & Trembr & 1.668 & 23.76 & 2.470 & 1.315 & 22.63 & 1.948 \\
    TrGNN & 1.255 & 15.89 & 2.415 & 1.559 & 17.68 & 2.763 & PIM   & 1.692 & 24.65 & 2.487 & 1.322 & 23.67 & 1.920 \\
    STGODE & 1.391 & 17.34 & 2.297 & 1.628 & 18.76 & 2.602 & Toast & 1.720 & 22.78 & 2.582 & 1.331 & 21.96 & 1.976 \\
    ST-Norm & 1.270 & 15.64 & 2.276 & 1.485 & 17.00 & 2.593 & JCLRNT & 1.799 & 24.91 & 2.576 & 1.370 & 23.95 & 1.987 \\
    SSTBAN & \underline{1.175} & \underline{14.11} & 2.193 & \underline{1.454} & 16.90 & \underline{2.491} & START & \underline{1.522} & \underline{22.26} & \underline{2.169} & \underline{1.182} & \underline{20.02} & \underline{1.749} \\
    \midrule
    TRACK & \textbf{1.094} & \textbf{13.32} & \textbf{2.141} & \textbf{1.363} & \textbf{15.42} & \textbf{2.471} & TRACK & \textbf{1.426} & \textbf{20.74} & \textbf{1.988} & \textbf{1.143} & \textbf{19.43} & \textbf{1.612} \\
    \bottomrule
    \end{tabular}%
  }
  \caption{Performance Comparsion.}
  \label{tab:performance}%
\end{table*}%

We utilize two real-world datasets from two cities in China, \ie Xi'an and Chengdu, collected from the DiDi GAIA project in October and November 2018. The statistics for the two datasets are presented in Table~\ref{tab:data_des}. The duration of each time slice is 30 minutes. We chronologically split the traffic state data into a training, validation, and testing set with a ratio of 6:2:2.
All experiments are repeated 10 times and the average results are reported according to Student’s t-test at the 0.01 significance level. The number of dimension $d$ in the representation learning is searched over $\{32, 64, 128\}$. The numbers of traffic transformer encoder layers, trajectory transformer encoder layers and co-attentional transformer encoder layers are 3, 6 and 2 respectively. More details are available at the code repository.

We evaluate the learned segment representations and trajectory representations on two downstream tasks, respectively, \ie Multi-Step Traffic State Prediction (MSTSP) and Travel Time Estimation (TTE). For the MSTSP task, we compare \name with 7 traffic state prediction methods, including DCRNN~\cite{DCRNN}, GWNET~\cite{GWNET}, MTGNN~\cite{MTGNN}, TrGNN~\cite{TrGNN}, STGODE~\cite{STGODE}, ST-Norm~\cite{ST-Norm} and SSTBAN~\cite{SSTBAN}.
For the TTE task, we compare \name with 7 trajectory representation learning methods, including traj2vec~\cite{trajectory2vec}, t2vec~\cite{t2vec}, Trembr~\cite{Trembr}, PIM~\cite{PIM}, Toast~\cite{Toast}, JCLRNT~\cite{JCLRNT} and START~\cite{START}.

\subsection{Performance Comparison}

Table~\ref{tab:performance} reports the results for two downstream tasks. The bold results are the best, and the underlined results are the second best. It can be seen that our proposed \name model outperforms all baselines on both datasets. This demonstrates the effectiveness of \name in learning effective road network representations and trajectory representations.

For the MSTSP task, MTGNN and SSTBAN achieve competitive performance compared to other baselines. This is because MTGNN proposes an adaptive graph generation module to reflect realistic spatial correlations while SSTBAN employs a self-supervised learner and designs a spatial botteleneck attention mechanism to capture global spatial dynamics. In contrast, \name achieves better performance because it further incorporates transition patterns of segments based on trajectory data in modeling the spatial-temporal dynamics. For the TTE task, START consistently outperforms other baselines for all metrics and datasets. One important reason is that it captures the temporal dynamics of trajectories and therefore enables the trajectories passing through the same route at different time slices may have different representations. However, START only encodes the timestamp information of trajectories for the periodic patterns of urban traffic, while \name also learns the spatial-temporal dynamics in the short-term traffic states.

\subsection{Ablation Study}
To further investigate the effects of different components in \name, we perform ablation studies with the following variants. (1) \textit{w/o Traf}: this variant eliminates modeling of the spatial-temporal dynamics in traffic state sequences; (2) \textit{w/o Traj}: this variant eliminates the process of trajectory representation learning and modeling of dynamic spatial features in trajectories; (3) \textit{w/o Match}: this variant eliminates the trajectory-traffic state matching task; (4) \textit{w/o TMask}: this variant removes the loss $\mathcal{L}^{Traj}_T$ of the MTP task, which means no masking prediction for timestamps; (5) \textit{w/o SMask}: this variant removes the loss $\mathcal{L}^{Traj}_S$ of the MTP task, which means no masking prediction for segments; (6) \textit{w/o Contra}: this variant removes the loss $\mathcal{L}^{Traj}_{Con}$ of the CTL task.

\begin{figure}[bt!]
    \centering
    \includegraphics[width=0.95\columnwidth]{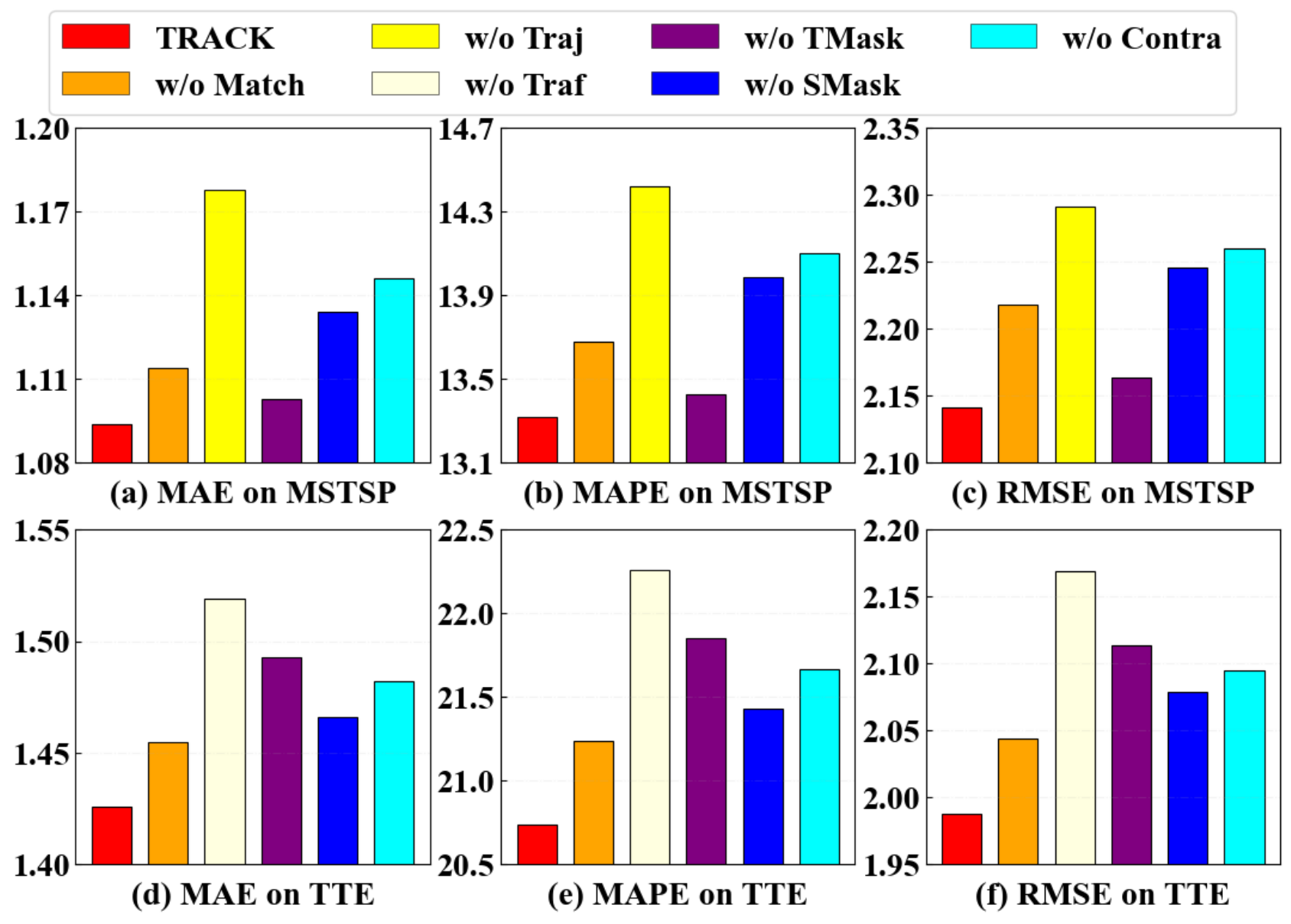}
    \caption{Ablation Study on the Xi'an Dataset.}
    \label{fig:abla}
\end{figure}

Figure~\ref{fig:abla} shows the performance of these variants on the MSTSP and TTE tasks of the Xi'an dataset. The following observations can be made. Firstly, the variants \textit{w/o Traj} and \textit{w/o Traf} are consistently inferior to \name on both tasks. This demonstrates that traffic state and trajectory data can serve as side information of each other to enhance the representation learning process of road networks and trajectories. Specifically, trajectory representations can perceive dynamic traffic states on the road network, while segment representations can perceive dynamic dependencies between upstream and downstream traffic flows. Secondly, \name performs better than \textit{w/o Match}. This indicates that the Trajectory-Traffic State Matching task can indeed help to enhance the representation learning process of the whole model. Third, the performance of \textit{w/o TMask}, \textit{w/o SMask}, and \textit{w/o Contra} on both tasks is inferior to TRACK. This indicates that both MTP and CTL tasks contribute to more accurate representations of dynamic road networks and trajectories, which in turn impacts their performance on two downstream tasks.

\subsection{Case Study}
In this section, we conduct two case studies to visualize and analyze the dynamic segment representations and trajectory representations, which enhance \name's interpretability.

\begin{figure}[bt!]
    \centering
    \includegraphics[width=\columnwidth]{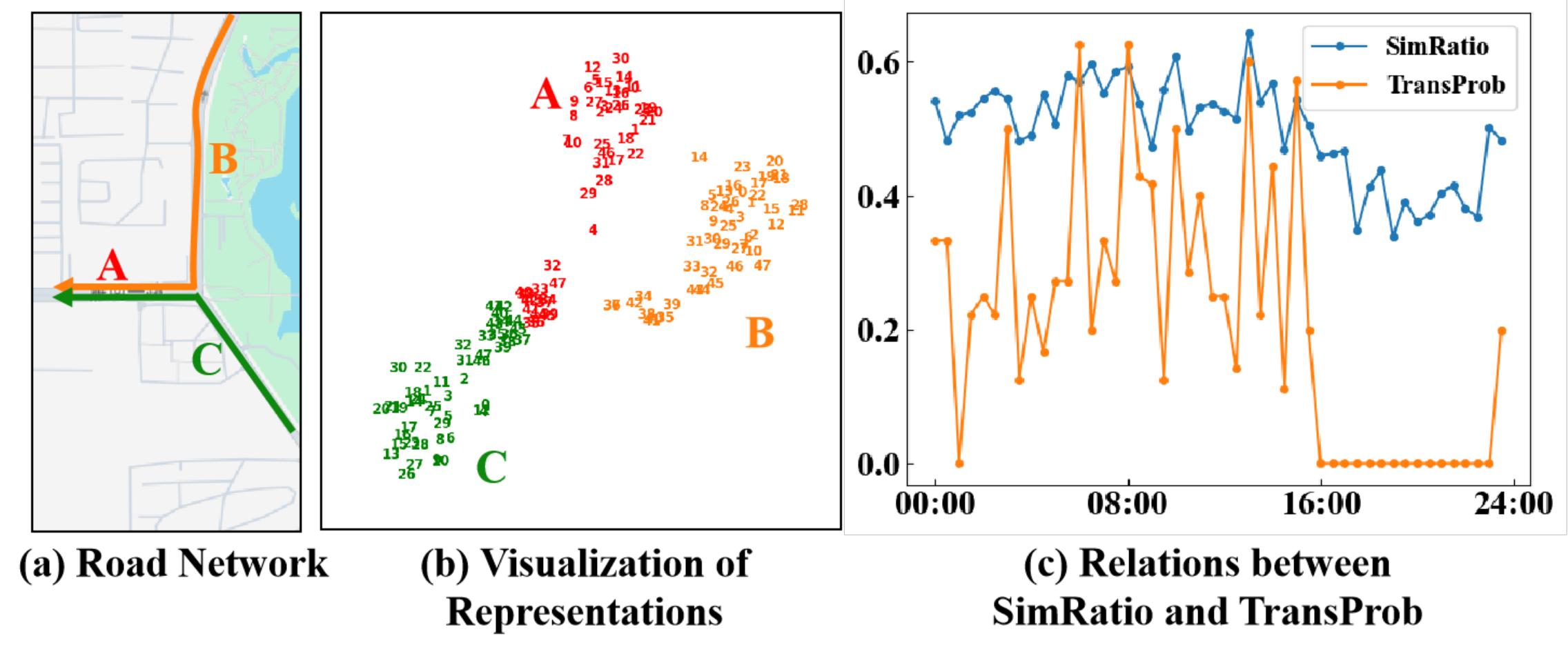}
    \caption{Case Study of Dynamic Segment Representations.}
    \label{fig:case1}
\end{figure}

\paratitle{Case 1: Study on Dynamic Road Segment Representations.}
One advantage of the proposed model is to learn dynamic road segment representations. We take segments A, B and C in Figure~\ref{fig:case1}(a) as an example to investigate this type of representation. The incoming traffic flow of segment A is composed of the outgoing traffic flow of segments B and C. We take October 11th as the observation period and set the window of the time slice as 30 minutes. Then we adopt t-SNE~\cite{tSNE} to visualize the learned segment representations in Figure~\ref{fig:case1}(b), where each point corresponds to the representation of a segment within a time slice. We can see that the learned segment representation indeed changes over time. Moreover, we use \textit{SimRatio(A,B)} to measure the ratio of the similarity between segments A and B and the similarity between segments A and C.
We then plot \textit{SimRatio(A,B)} and the transition probabilities from segment B to A, denoted as \textit{TransProb(B$\rightarrow$A)}, over time in Figure~\ref{fig:case1}(c). It can be observed that the transition patterns indeed change over time. More importantly, we can see that from 16:00 to 23:00 there is a significant decrease in \textit{SimRatio(A,B)}, which is due to the fact that there were no trajectories from B to A in that period. This suggests that the learned segment representations can indeed capture the dynamic transition patterns between road segments.

\begin{figure}[bt!]
    \centering
    \includegraphics[width=\columnwidth]{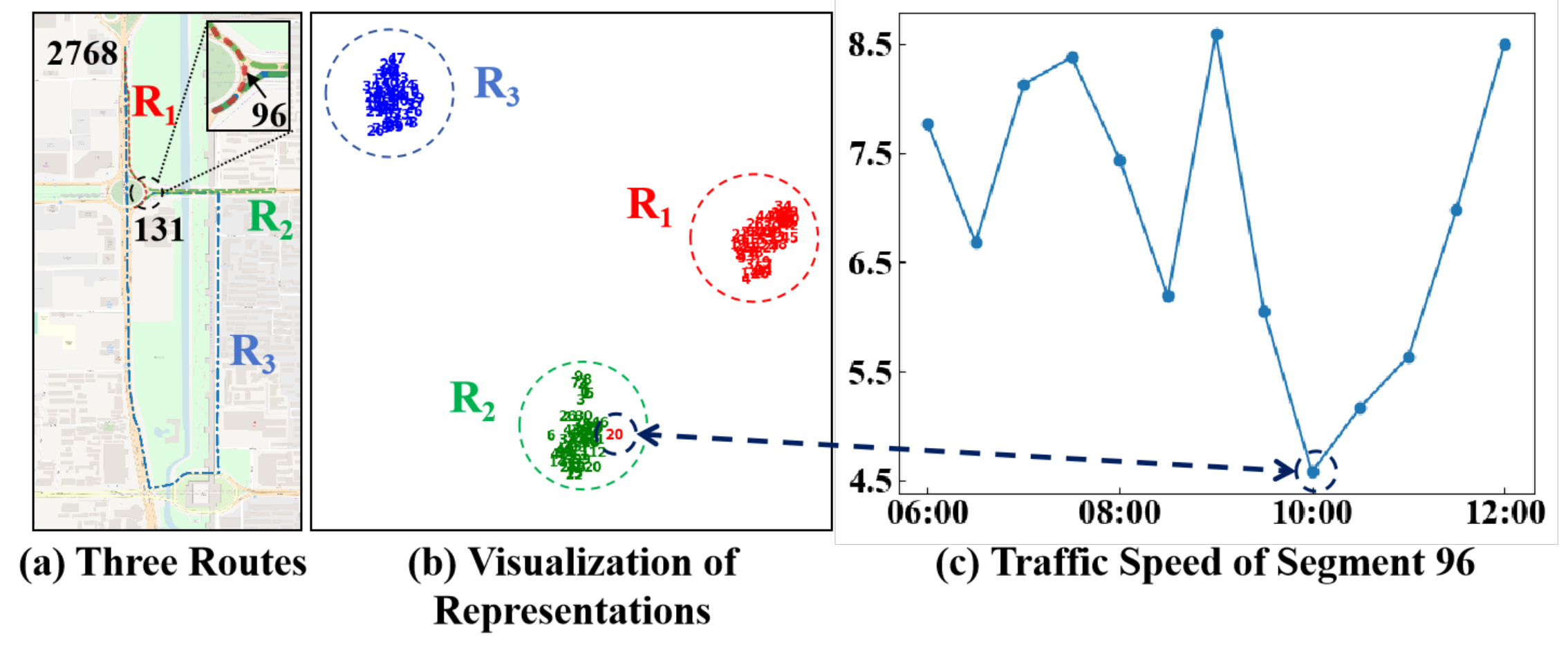}
    \caption{Case Study of Trajectory Representations.}
    \label{fig:case2}
\end{figure}

\paratitle{Case 2: Study on Dynamic Trajectory Representations.} 
Another advantage of the proposed model is that with the information exchange of different views, the learned trajectory representations can also be influenced by the dynamics of traffic states. 
We take the three routes $R_1$, $R_2$, and $R_3$ from segment 131 to segment 2768 on November 1st shown in Figure~\ref{fig:case2}(a) as an example. In Figure~\ref{fig:case2}(b), we visualize trajectory representations of the three routes at different time slices, which shows that the learned trajectory representations of the same route are indeed dynamic over time. Moreover, compared to $R_3$, $R_1$ and $R_2$ are closer in the trajectory representation space. This can be explained by the road network's spatial semantics because $R_2$ has a shorter detouring path than $R_3$ and is more similar to $R_1$. Meanwhile, the trajectory representation of $R_1$ at the 20th time slice (\ie departing at 10:00) is more similar to those of $R_2$ than those of $R_1$ at other time slices. This can be explained by that there was a sharp decrease in traffic speed on segment 96 at 10:00 as shown in Figure~\ref{fig:case2}(c), which makes the semantics of $R_1$ at that time slice resembles the semantics of detouring in $R_1$. This indicates that trajectory representation generated by \name can indeed perceive the dynamic traffic states of the road segments visited in the trajectory.

\section{Related Work}

\textbf{Road Network Representation Learning} aims to transform the road network into a general low-dimensional representation matrix. Since graph representation learning methods can model the topological structure of the road network~\cite{DeepWalk,LINE,node2vec,GraphSAGE}, some existing methods~\cite{IRN2Vec,chen,SARN} take the spatial correlations of road segments into account by using graph representation learning methods. 
A more recent method~\cite{HRNR, yu2024bigcity} takes the transition patterns into account when modeling the road networks. In summary, our proposed method is the first attempt to jointly model the dynamics of road segments based on trajectory and traffic state data.

\textbf{Trajectory Representation Learning} aims to transform the trajectory into a general low-dimensional representation vector. Early TRL studies~\cite{trajectory2vec,t2vec} obtain trajectory representations through the reconstruction task. Recent TRL methods~\cite{TMRN,PIM,CSTRM,JCLRNT,WSCCL,Toast,MMTEC,START} primarily first obtain the road network representations and then derive trajectory representations through sequential models with self-supervised tasks. The trajectory representation is assumed to be static in most methods, with only a few encoding temporal information in trajectories. For example, Trembr~\cite{Trembr} and START~\cite{START} reconstructs timestamps during the decoding process. However, they do not consider the impact of dynamic traffic states on trajectory representations, which is one of our main contributions.

\section{Conclusion}
We propose \name, a novel dynamic road network and trajectory representation learning framework, to jointly model traffic state data and trajectory data. Extensive experiments on two real-world datasets showcase the performance of \name and provide interpretations of dynamic road segment and trajectory representations. 
\name offers a promising way for improving traffic-related tasks, contributing to more efficient and sustainable urban management.

\section{Acknowledgments}
Prof. Jingyuan Wang's work was partially supported by the National Natural Science Foundation of China (No. 72171013, 72222022), and the Special Fund for Health Development Research of Beijing (2024-2G-30121). Prof. Junjie Wu’s work was partially supported by the National Natural Science Foundation of China (72242101, 72031001), and the Outstanding Young Scientist Program of Beijing Universities (JWZQ20240201002). Dr. Hao Lin's work was partially supported by the National Natural Science Foundation of China (72301017) and the Shenzhen Science and Technology Program (CJGJZD20230724093201004).
\bibliography{aaai25}

\end{document}